\documentclass[letterpaper]{article} % DO NOT CHANGE THIS
\usepackage{aaai25}  % DO NOT CHANGE THIS
\usepackage{times}  % DO NOT CHANGE THIS
\usepackage{helvet}  % DO NOT CHANGE THIS
\usepackage{courier}  % DO NOT CHANGE THIS
\usepackage[hyphens]{url}  % DO NOT CHANGE THIS
\usepackage{graphicx} % DO NOT CHANGE THIS
\usepackage{natbib}  % DO NOT CHANGE THIS AND DO NOT ADD ANY OPTIONS TO IT
\usepackage{caption} % DO NOT CHANGE THIS AND DO NOT ADD ANY OPTIONS TO IT
\usepackage{algorithm}
\usepackage{newfloat}
\usepackage{listings}

\usepackage{multirow}
\usepackage{multicol}
\usepackage{booktabs}
\usepackage{amsmath}
\usepackage{amsfonts}
\usepackage{algpseudocode}
\newcommand{\beq}[1]{\begin{equation}#1\end{equation}}

\usepackage[table,xcdraw]{xcolor}

\DeclareCaptionStyle{ruled}{labelfont=normalfont,labelsep=colon,strut=off} % DO NOT CHANGE THIS
\floatstyle{ruled}
\newfloat{listing}{tb}{lst}{}
\floatname{listing}{Listing}
\title{SQLFixAgent: Towards Semantic-Accurate Text-to-SQL Parsing \\via Consistency-Enhanced Multi-Agent Collaboration}
\author {
	Jipeng Cen\textsuperscript{\rm 1},
	Jiaxin Liu\textsuperscript{\rm 4},
	Zhixu Li\textsuperscript{\rm 2,\rm 3},
	Jingjing Wang\textsuperscript{\rm 1}\thanks{Jingjing Wang is the corresponding author.}
}
\affiliations {
	\textsuperscript{\rm 1}School of Computer Science \& Technology, Soochow University, Suzhou, China\\
	\textsuperscript{\rm 2}School of Information, Renmin University of China, Beijing, China\\
	\textsuperscript{\rm 3}International College (Suzhou Research Institute), Renmin University of China, Suzhou, China\\
	\textsuperscript{\rm 4}iFLYTEK Research (Suzhou), China\\
	jpcen@stu.suda.edu.cn, jxliu6@iflytek.com, zhixuli@ruc.edu.cn, djingwang@suda.edu.cn
}

\begin{document}

\maketitle

\begin{abstract}
	While fine-tuned large language models (LLMs) excel in generating grammatically valid SQL in Text-to-SQL parsing, they often struggle to ensure semantic accuracy in queries, leading to user confusion and diminished system usability. To tackle this challenge, we introduce SQLFixAgent, a new consistency-enhanced multi-agent collaborative framework designed for detecting and repairing erroneous SQL. Our framework comprises a core agent, SQLRefiner, alongside two auxiliary agents: SQLReviewer and QueryCrafter. The SQLReviewer agent employs the rubber duck debugging method to identify potential semantic mismatches between SQL and user query. If the error is detected, the QueryCrafter agent generates multiple SQL as candidate repairs using a fine-tuned SQLTool. Subsequently, leveraging similar repair retrieval and failure memory reflection, the SQLRefiner agent selects the most fitting SQL statement from the candidates as the final repair. We evaluated our proposed framework on five Text-to-SQL benchmarks. The experimental results show that our method consistently enhances the performance of the baseline model, specifically achieving an execution accuracy improvement of over 3\% on the Bird benchmark. Our framework also has a higher token efficiency compared to other advanced methods, making it more competitive.
	
\end{abstract}

\section{Introduction}

Traditional data querying methods require users to have expertise in Structured Query Language (SQL), posing a significant barrier for non-technical users. Text-to-SQL parsing aims to overcome this obstacle by automatically converting natural language questions into SQL queries ~\citep{qin2022survey, deng2022recent}. 

Previous Text-to-SQL methods mainly rely on pretrained models to encode the input sequence ~\citep{wang2019rat,hui-etal-2022-s2sql}, then decode the SQL query using abstract syntax trees (AST)~\citep{guo-etal-2019-towards,wang2019rat,cao2023astormer}. However, recent advancements in decoder-based LLMs have transformed the field of NLP. The researchers are actively exploring the potential of LLMs in Text-to-SQL parsing. These methods can be divided into two categories: Fine-tuning LLMs and Prompting LLMs. 
\begin{figure}[tb]
	\centering
	\small
	\includegraphics[width=0.98\columnwidth]{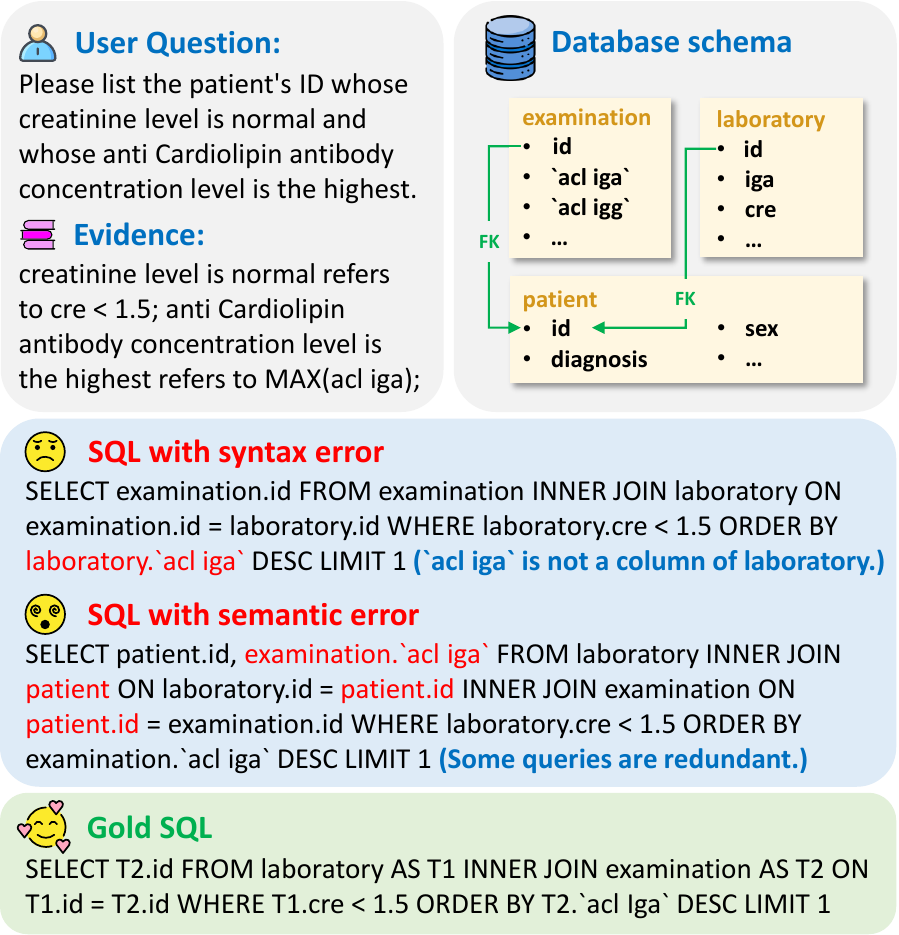}
	\caption{A complex example of Text-to-SQL from Bird. We demonstrate two common types of errors happened in LLM-based parsing: \textbf{SQL with syntax errors}, where missing table names are easily detected, while \textbf{SQL with semantic errors} is often executed, leading to user confusion. }
	\label{fig:example}
\end{figure}
The former fine-tunes the LLMs on Text-to-SQL downstream tasks, aiming to enhance their capabilities by high-quality data~\citep{li2024codes,sun2024sqlpalm}. The latter prompts the proprietary LLMs with hundreds of billions parameters, such as GPT-4~\citep{openai2024gpt4}, to push the boundaries of Text-to-SQL by leveraging their superior In-Context-Learning capabilities~\citep{gao2023texttosql,wang2024macsql,xie2024decomposition}. However, it is worth noting that due to the concerns of inference overheads and data privacy, Fine-tuning LLMs remains the preferred choice for most practical applications. With the improvement of open source LLMs~\citep{touvron2023llama,li2023starcoder}, these methods achieved commendable performance on some simpler Text-to-SQL benchmarks~\citep{yu2019spider,yu-etal-2019-sparc}, comparable to those state-of-the-art prompting-based methods. 
\begin{figure*}[tb]
	\centering
	\small
	\includegraphics[width=0.9\textwidth]{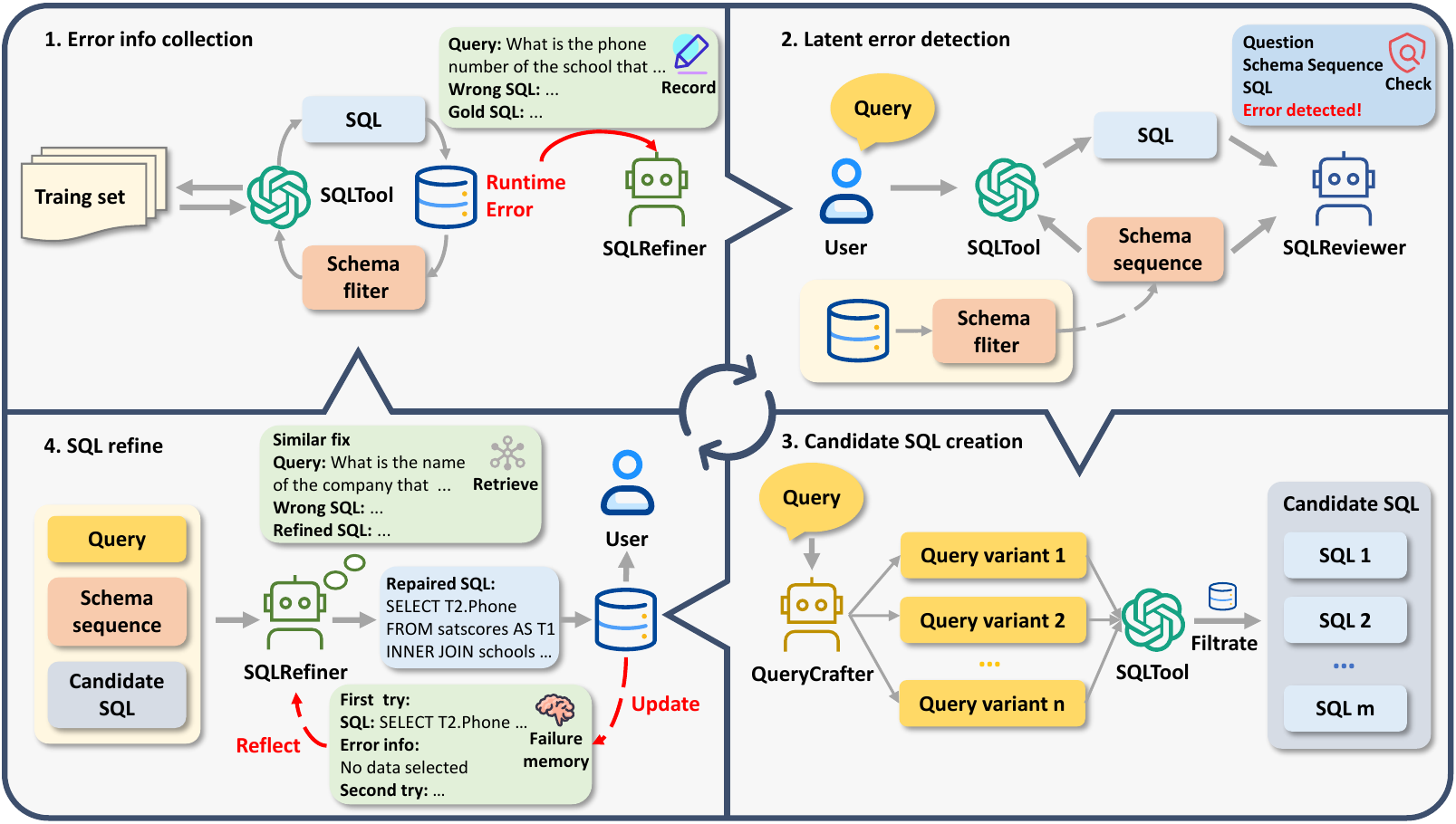}
	\caption{The overview of SQLFixAgent framework, which comprises three agents: (i) the \textit{SQLReviewer}, which detects syntax and semantic errors in SQL generated by SQLTool based on user query and schema sequence, (ii) the \textit{QueryCrafter}, which generates multiple variants of user query, then utilizes SQLTool to generate multiple SQL as candidate repairs, and (iii) the \textit{SQLRefiner}, which records runtime errors from SQLTool, and selects the optimal SQL as final repair from candidates.}
	\label{fig:overview}
\end{figure*}

Although the fine-tuned LLMs excel in generating syntactically valid SQL statements, they often struggle to produce semantically consistent ones in real-world scenarios~\citep{sun2023battle}. Given user query and the database schema, Figure \ref{fig:example} illustrates two types of errors in LLM-generated SQL: syntax errors and semantic errors. Compared with syntax errors, semantic errors are more difficult to detect since they can be executed smoothly in the database. These semantic errors lead to confusing execution results, significantly reducing the reliability of the Text-to-SQL system. To enhance the capability of Text-to-SQL system in generating semantically accurate results, we introduce SQLFixAgent, a consistency-enhanced multi-agent collaborative framework designed to detect and correct errors in SQL generated by LLMs. Our framework comprises a core \textit{SQLRefiner} agent for repair decision, accompanied by two auxiliary agents, \textit{SQLReviewer} and \textit{QueryCrafter}, for latent error detection and candidate SQL generation. Specifically, the SQLReviewer employs the Rubber Duck Debugging method to identify potential semantic discrepancies between SQL and user query, then initiate the repair request to other agents. Upon receiving the request, the QueryCrafter utilizes SQLTool to generates multiple candidate SQL, then, the SQLRefiner selects the one that best aligns with the query as the final repair from the candidates by retrieving similar repair examples and reflecting from failure memory.

In our experiments, we use the fine-tuned Codes~\citep{li2024codes} as the SQLTool used by agents for Text-to-SQL parsing and employ GPT-3.5-turbo as the backbone LLM for three agents. We evaluate our framework on five Text-to-SQL benchmarks: Bird~\citep{li2023llm}, Spider~\citep{yu2019spider} and its three robustness variants~\citep{gan2021exploring,gan2021robustness,Deng_2021}. The experimental results demonstrate that our SQLFixagent can effectively reduce the syntactic and semantic errors generated by SQLTool. On this basis, we conduct comprehensive ablation experiments on the framework. We compare the fine-grained capabilities of the backbone LLMs and evaluate the agents' performance on sub-tasks, further revealing how the foundation LLMs serve as agents to assist the fine-tuned LLM in generating more semantic-accurate SQL queries. Moreover, to achieve a practical Text-to-SQL solution, we compared the token efficiency of our approach with other advanced methods, demonstrating its competitive performance.

\paragraph{Contributions} (1) We highlight the semantic inconsistency error between the LLM-generated SQL and the user's query in Text-to-SQL parsing. To address this, we propose SQLFixAgent, a consistency-enhanced multi-agent collaborative framework designed to detect and repair the errors in LLM-generated SQL, including syntax and semantic errors. (2) We conduct extensive evaluation and analysis on five Text-to-SQL benchmarks, demonstrating that our framework consistently enhances the performance of the baseline model, specifically achieving an execution accuracy improvement of over 3\% on the Bird benchmark. Additionally, our approach achieves higher token efficiency compared to other advanced methods, enhancing its competitiveness.

\section{Related Work}
\subsection{Text-to-SQL with LLMs}
\paragraph{Prompting LLMs} Recent advancements in large language models (LLMs) have resulted in groundbreaking capabilities~\citep{openai2024gpt4,geminiteam2024gemini}. To explore the potential of foundation LLMs in Text-to-SQL parsing, advanced prompting approaches tailored to the task have been proposed. DAIL-SQL~\citep{gao2023texttosql} systematically examined prompt engineering for LLM-based Text-to-SQL methods, it employed dynamic
few-shot examples by considering the similarity of
both the user questions and the SQL queries. Subsequent studies introduced frameworks like DIN-SQL~\citep{pourreza2023dinsql}, C3-SQL~\citep{dong2023c3}, and StructGPT~\citep{jiang2023structgpt}, which aim to verify user queries, simplify databases representation, generate and integrate answers by utilizing techniques such as self-consistency~\citep{wei2022chain}, chains of thought~\citep{NEURIPS2022_9d560961}, and least-to-most
prompting~\citep{zhou2023leasttomost}. However, they overly rely on the capabilities of proprietary LLMs with hundreds of billions parameters, raising concerns of data privacy and inference overheads.

\paragraph{Fine-tuning LLMs} Compared to prompting LLMs, fine-tuning LLMs is more widely adopted in enterprises. DAIL-SQL~\citep{gao2023texttosql} has investigated fine-tuning open-source LLMs. However, due to the smaller sizes, their performance remains lower than prompting larger LLMs such as GPT-4. Some research is dedicated to improving the capabilities of fine-tuned LLMs. CodeS~\citep{li2024codes} applied an incremental pre-training technique to train the Text-to-SQL LLMs on a meticulously constructed pre-training corpus, and achieves impressive results. SQL-PaLM~\citep{sun2024sqlpalm} focuses on larger-scale LLMs, investigating the potential for significant gains with increased model size due to the emergent capabilities of LLMs. However, although the fine-tuned LLMs excel at generating syntactically valid SQL statements, they often struggle to produce semantically accurate queries, which is the challenge we need to address.

\subsection{LLM-based Agents}
Following the rise of large language models, LLM-based agents have become one of the most prevalent AI systems~\citep{Wang_2024,xi2023rise,durante2024agent}. These agents learn to interact with the external world through action-observation pairs articulated in natural language. AutoGPT~\citep{autogpt2023} is an early open-source implementation of the AI agent, following a single-agent paradigm. It enhances the AI model with a number of useful tools but lacks support for multi-agent collaboration. In the programming domain, MetaGPT~\citep{hong2023metagpt} innovatively simulates the operational structure of a traditional software company. This framework drives the GPT agents to collaborate on user-defined programming tasks by assigning them to different roles: product manager, project manager, or engineer. Similar work includes FixAgent~\citep{lee2024unified}, an automated and integrated debugging framework that achieves its goals through the collaboration of multiple agents. In the Text-to-SQL field, MAC-SQL~\citep{wang2024macsql} introduced a pioneering multi-agent framework, enabling multiple LLM-based agents to collaboratively tackle Text-to-SQL parsing tasks and effectively manage the complexity and diversity encountered in real-world query scenarios. However, the application of LLM-based agents to detect and fix SQL errors generated by fined-tuned LLMs remains under-explored, which is our primary focus.

\section{Problem Formulation}
\paragraph{Text-to-SQL Task} Formally, given a natural language question $\mathcal{Q}$ based on a database schema $\mathcal{S}$ and a knowledge evidence $\mathcal{K}$, Text-to-SQL parsing aims to translate $\mathcal{Q}$ into a SQL query $y$ that can be executed on relational database $\mathcal{D}$ to answer the question $\mathcal{Q}$. The database schema $\mathcal{S}$ of database $\mathcal{D}$ includes (1) a set of tables $\mathcal{T} =\left\{ \mathcal{T} _1,\mathcal{T} _2,...,\mathcal{T} _m \right\}$, (2) a set of columns $\mathcal{C} =\left\{ \mathcal{C} _{\mathcal{T} _1}^{1},...,\mathcal{C} _{\mathcal{T} _1}^{n},\mathcal{C} _{\mathcal{T} _2}^{1},...,\mathcal{C} _{\mathcal{T} _2}^{n},\mathcal{C} _{\mathcal{T} _m}^{1},...,\mathcal{C} _{\mathcal{T} _m}^{n} \right\}$ associated with the tables, (3) a set of foreign key relations $\mathcal{R} =\left\{ \left( \mathcal{C} _{k}^{i},\mathcal{C} _{h}^{j} \right) | \mathcal{C} _{k}^{i},\mathcal{C} _{h}^{j}\in \mathcal{C} \right\}$. Here, $m$ and $n$ denote the number of table names and column names, respectively. With the language model policy $\theta $, the Text-to-SQL parsing task could be formulated as:
\beq{
	y=f\left( \mathcal{Q} ,\mathcal{S} ,\mathcal{K}|\theta  \right) ,
}
\paragraph{SQL Repair Task} We consider the fine-tuned LLM as SQLTool in SQL Repair task, and use SQLFixAgent to detect and repair the erroneous results generated by SQLTool.
\begin{algorithm}[H]
	\caption{The algorithm of SQLFixAgent}\label{alg:sqlfixagent}
	\begin{algorithmic}[1]
		\Require \texttt{question q}, \texttt{database db}, \texttt{knowledge kg}, \texttt{training set t}
		\Ensure \texttt{label}, \texttt{sql}
		\State // Collect runtime error from $\mathrm{SQLTool}$
		\State \textcolor[HTML]{3078BE}{$\mathrm{SQLRefiner_{init}}$}($\mathrm{SQLTool}$, \texttt{t})
		\State \texttt{db} = \textcolor[HTML]{3078BE}{$\mathrm{SchemaFilter}$}(\texttt{q}, \texttt{db}, \texttt{kg})
		\State \texttt{psql} = \textcolor[HTML]{3078BE}{$\mathrm{SQLTool}$}(\texttt{q}, \texttt{db}, \texttt{kg})
		\State // Detect syntax errors and semantic error in \texttt{psql}
		\State \texttt{label} = \textcolor[HTML]{3078BE}{$\mathrm{SQLReviewer}$}(\texttt{psql}, \texttt{db}, \texttt{kg})
		
		\If{label}
		\State \texttt{sql = psql}
		\Else
		\State \texttt{count} = \texttt{0}
		\While{$\texttt{count} < \texttt{maxTryTimes}$}
		\State // Generate candidate SQL
		\State \texttt{sqls} = \textcolor[HTML]{3078BE}{$\mathrm{QueryCrafter}$}(\texttt{q}, \texttt{db}, \texttt{kg}, $\mathrm{SQLTool}$)
		\State // Retrieve similar repair
		\State \texttt{eg} =
		\textcolor[HTML]{3078BE}{$\mathrm{SQLRefiner_{retrieve}}$}($\mathrm{SQLTool}$, \texttt{t})
		\State \texttt{sql} = \textcolor[HTML]{3078BE}{$\mathrm{SQLRefiner}$}(\texttt{q}, \texttt{db}, \texttt{kg}, \texttt{sqls}, \texttt{eg})
		\State \texttt{pass}, \texttt{err} = \textcolor[HTML]{3078BE}{$\mathrm{executeAndAnalyze}$}(\texttt{sql}, \texttt{db})
		\If{\texttt{pass}}
		\State \textbf{break}
		\Else
		\State // Refresh failure memory
		\State \textcolor[HTML]{3078BE}{$\mathrm{SQLRefiner_{refresh}}$}(\texttt{sql}, \texttt{err})
		\EndIf
		\EndWhile
		\EndIf
		\State \Return  \texttt{label}, \texttt{sql}	
	\end{algorithmic}
\end{algorithm}

\section{Method}
\subsection{Overview}
We introduce SQLFixAgent, a consistency-enhanced multi-agent collaborative framework designed to detect and repair the syntax and semantic errors in SQL generated by SQLTool. As illustrated in Figure~\ref{fig:overview}, our framework comprises a core \textit{SQLRefiner} agent for generating the final repaired SQL, accompanied by two auxiliary agents, the \textit{SQLReviewer} and the \textit{QueryCrafter}, for syntactic and semantic error detection and candidate repair generation. Specifically, the SQLReviewer employs the Rubber Duck Debugging method to check whether the SQL aligns with the user's query intent. If the error is detected, the SQLRefiner and QueryCrafter is tasked with making the SQL correction. Specifically, The QueryCrafter utilizes SQLTool to generates multiple candidate SQL statements and the SQLRefiner selects the most appropriate one as the final repair by retrieving similar repair and reflecting from failure memory.

\subsection{SQLReviewer} 
LLM-based Text-to-SQL parsing typically encounters two types of errors: syntax errors and semantic errors. The former can usually be detected and intercepted through the pre-execution check in database. In contrast, the latter can be executed smoothly, producing confusing results for the user and reducing the reliability of system. SQLReviewer is designed to detect the potential errors in SQL statements generated by LLMs. We provide SQLReviewer with a real environment connected to the database to capture syntax errors in SQL. Then, for those pass the syntax inspection, we further check whether their semantics align with the user's query based on the relevant database schema and evidence.

\begin{figure}[tb]
	\centering
	\small
	\includegraphics[width=0.98\columnwidth]{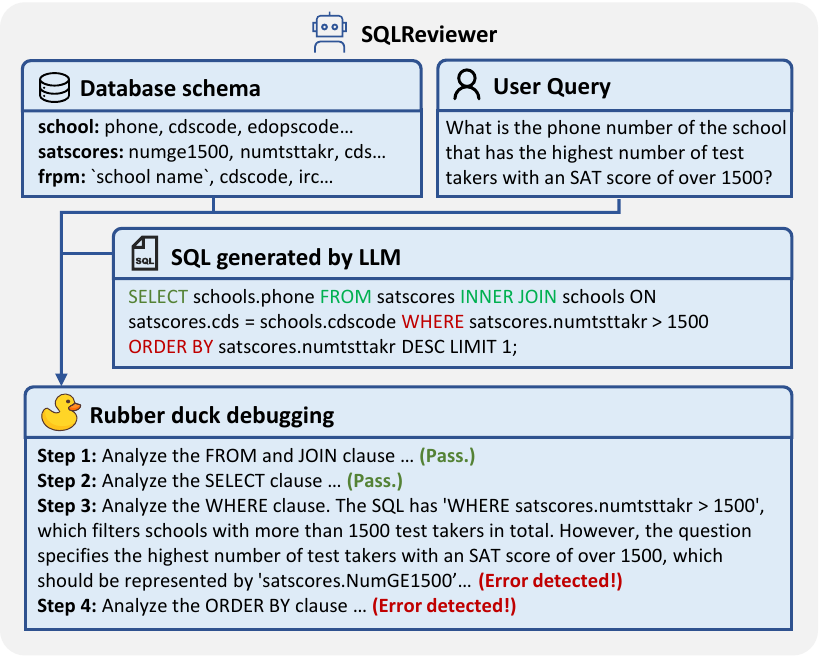}
	\caption{The SQLReviewer Agent Illustration.}
	\label{fig:SQLReviewer}
\end{figure}

Based on the opinion: \textit{LLMs closely mimic developers when performing coding-related tasks, so they can benefit from general software engineering principles}~\citep{lee2024unified}. We adopt the principle of rubber duck debugging in the design of SQLReviewer, a method where a programmer explains his code line-by-line to an inanimate object (like a rubber duck) to help identify and understand errors, as shown in the Figure~\ref{fig:SQLReviewer}. We assign SQLReviewer the role of a database administrator and require it to step-by-step verify whether each part of the SQL statement aligns with the user's query intent, thereby detecting potential semantic mismatch errors. If no error is detected, the original SQL will be returned as the final result. Otherwise, the message will be passed to other agents for  further processing.

\subsection{QueryCrafter} 
Although the fine-tuned LLMs excels at generating syntactically correct SQL statements, it remains challenging to produce ones that completely align with the user's query intent~\citep{sun2023battle}. To mitigate this challenge, we designed QueryCrafter, which leverages SQLTool to generate multiple candidate SQL for SQLRefiner to select the most semantic-accurate one as the final repair. Given a user query $\mathcal{Q}$, QueryCrafter generates a list of variants $\mathcal{Q}^{'}=\left\{ \mathcal{Q}_{1}^{'},\mathcal{Q}_{2}^{'},...,\mathcal{Q}_{n}^{'} \right\}$ with similar length and equivalent semantics. Then, by using SQLTool, QueryCrafter generates multiple SQL statements $y=\left\{ y_1,y_2,...,y_m \right\}$ based on query variants $\mathcal{Q}^{'}$, the database schema $\mathcal{S}$ and the evidence $\mathcal{K}$. The candidate SQL statements $y_{candidate}=\left\{ y_1,y_2,...,y_n \right\}$, which will be passed to SQLRefiner, are obtained by further pre-execution filtering of the SQL statements \( y \) in the database \( \mathcal{D} \).

Some researches showed that employing beam search decoding for LLMs to generate multiple candidate SQL can effectively improve the Text-to-SQL accuracy~\citep{li2023resdsql,li2024codes}. However, the fine-tuned LLMs tend to overfit the training data, reducing the diversity of effective candidates. In contrast, QueryCrafter generates multiple equivalent user queries, introducing perturbations at the input of SQLTool to broaden the spectrum of generated candidates. 

\subsection{SQLRefiner} 
As the core of our SQLFixAgent framework, the primary function of SQLRefiner is to determine the final repair of SQL errors. Since it is powered by a foundation LLM that is not specifically trained for Text-to-SQL tasks, it lacks proficiency in directly generating accurate SQL statements but excels in semantic analysis and comprehension. Therefore, we regard SQLRefiner as essential for ensuring the framework produces semantically correct SQL. Given the user query $\mathcal{Q}$, the evidence $\mathcal{K}$, and the database schema $\mathcal{S}$, SQLRefiner retrieves similar repairs from the SQLTool error record as correction examples. Then, it selects the SQL $y$ that best matches the user query from the candidate SQL $y_{candidate}$ provided by QueryCrafter as the final repair, along with explanation. In some cases where QueryCrafter can not provide effective SQL candidates, SQLRefiner will attempt to correct the SQL statement on its own.

\begin{figure}[tb]
	\centering
	\small
	\includegraphics[width=0.98\columnwidth]{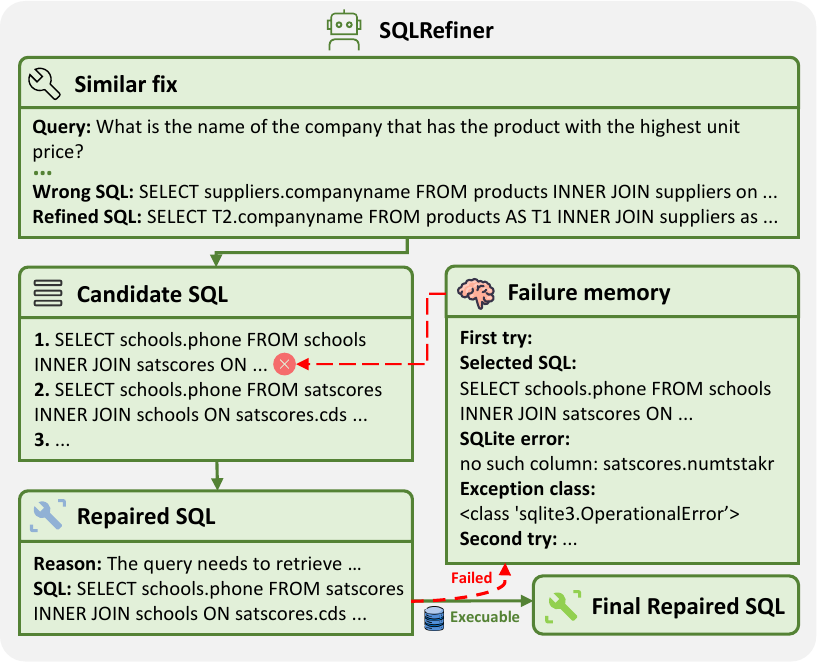}
	\caption{The SQLRefiner Agent Illustration.}
	\label{fig:SQLRefiner}
\end{figure}
Despite some prompting tricks being used to guide the LLMs toward expected outcomes, achieving an error-free repair in a single attempt remains difficult due to the hallucinate of LLM~\citep{huang2023survey}. To mitigate this challenge, we introduced Reflexion~\citep{shinn2023reflexion} mechanism in SQL repair. As shown in Figure~\ref{fig:SQLRefiner}. Upon selection or generation of a SQL statement, following MAC-SQL~\citep{wang2024macsql}, the SQLRefiner first assess its syntactic correctness, execution feasibility, and the retrieval of non-empty results from the database. If the check passes, the SQL statement will be taken as the final repair; otherwise, it will be written into the failure memory along with the execution error feedback from database. In the next attempt, SQLRefiner reflects on the failure memory and repeats the repair process until the generated result passes the execution check or the maximum number of correction is reached.

The essence of the SQLFixAgent framework lies in the collaboration between a foundational LLM and a specialized LLM trained on Text-to-SQL task. We believe their capabilities are complementary: while the foundational LLM possesses strong ICL abilities, it is not adept at generating effective SQL statements directly. On the other hand, the Text-to-SQL LLM excels at generating syntactically valid SQL statements, but often struggles to produce ones with semantic consistency. In the SQLFixAgent framework, the correctness of SQL syntax is primarily ensured by SQLTool, while the consistency of SQL semantics is guaranteed through the collaboration of QueryCrafter and SQLRefiner.

\section{Experiments}
\subsection{Datasets}
We evaluated our framework on two primary Text-to-SQL benchmarks: Spider~\citep{yu2019spider} and Bird~\citep{li2023llm}. Moreover, we assessed the robustness on three variants of Spider: Spider-DK~\citep{gan2021exploring}, Spider-Syn~\citep{gan2021robustness}, and Spider-Realistic~\citep{Deng_2021}. For the training sets of Spider and Bird, we used them to simulate runtime error record.

\textbf{Spider} is widely used to evaluate Text-to-SQL parsers across various databases, requiring models to demonstrate their adaptability to unfamiliar database structures. It offers a training set comprising 8,659 samples, a development set with 1,034 samples, and a test set with 2,147 samples, encompassing 200 distinct databases and 138 domains.

\textbf{BIRD} released by Alibaba DAMO Academy, is a new benchmark for large-scale database grounded Text-to-SQL evaluation. It contains 95 large-scale databases and high-quality Text-SQL pairs, with a data storage volume of up to 33.4GB spanning 37 professional domains. Compared to Spider, it focuses on integrating external knowledge reasoning to bridge natural language questions and database content, and addresses new challenges related to SQL efficiency when handling large databases.

\textbf{Spider-DK, Spider-Syn, Spider-Realistic} are variants derived from the original Spider dataset, introducing robustness challenges through domain knowledge introduction, synonym substitution, and explicit mention removal, respectively. They are specifically crafted to resemble queries that might be posed by users in real-world scenarios.

\subsection{Evaluation Metrics}
We adopt execution accuracy, exact match accuracy, and valid efficiency score to evaluate the performance of our framework. \textit{Execution accuracy (EX)} is defined as the proportion of questions in the evaluation set where the execution result of both the predicted and ground-truth queries are identical. \textit{Exact match accuracy (EM)}, introduced by Test-Suites~\citep{ruiqi20}, evaluates each clause as a set and compares the predicted clauses with their corresponding ones in the reference query. A predicted SQL query is considered correct only if all of its components match the ground
truth. \textit{Valid efficiency score (VES)}, introduced by Bird~\citep{li2023llm}, is designed to measure the efficiency of valid SQL. It considers both the accuracy and efficiency for the model-generated query.

\subsection{Implementation Details}
In evaluation, we utilized fine-tuned version of CodeS~\citep{li2024codes} with 3b and 7b parameters as the SQLTool within the framework, and all agents were powered by GPT-3.5-turbo. For SQLTool inference, a beam search produces 4 SQL candidates, we select the first executable one for further checking by SQLFixAgent. If the error is detected, SQLFixAgent attempts to repair it up to 3 times. These hyper-parameters are tuned on validation set. All experiments were conducted on a server equipped with 1 × AMD EPYC 7352 CPU and 8 × NVIDIA RTX 3090 GPU. To facilitate the related research, all codes will be released via Github\footnote{https://github.com/Cen-Jipeng-SUDA/SQLFixAgent}.

\subsection{Main Results}
\paragraph{Results on BIRD} Table~\ref{tab:bird_result} presents the performance of the SQLFixAgent framework compared to other competitive methods on the current most challenging Text-to-SQL benchmark, BIRD.
We observe that CodeS-7B + SQLFixAgent achieves better performance than standalone CodeS-7B, indicating that SQLFixAgent effectively reduces semantic and syntactic errors in the SQL generated by the LLMs. Surprisingly, it outperforms DAIL-SQL by 5.41\% EX with backbone models (CodeS-7B and GPT-3.5 Turbo) that their parameters significantly fewer than GPT-4, highlighting the value of SQL repair study. Specifically, CodeS-7B equipped with SQLFixAgent demonstrated a 3.00\% increase in EX and a 4.35\% increase in VES, while CodeS-3B equipped with SQLFixAgent improved EX by 3.65\% and VES by 5.65\%, showing even greater improvement. This result is attributed to the weaker capability of CodeS-3b, which leads to more correctable  errors in the generated SQL.

\paragraph{Results on Spider and its Variants} As shown in Table~\ref{tab:spider_result}, SQLFixAgent enhances the
EX performance of the CodeS-3B and CodeS-7B baseline by 1.4\% and 0.8\% on Spider's dev set. Since SQLFixAgent is powered by GPT-3.5 Turbo without fine-tuning, it does not contribute to improving exact match accuracy. On the other hand, CodeS-7b with SQLFixAgent shows a 0.6\% EX improvement on test set. This represents the performance limit of GPT-3.5 Turbo on Spider. Across three robustness variants of Spider, as shown in Table~\ref{tab:variant_result}, our framework achieved CodeS-7B EX improvements of 0.8\%, 0.4\%, and 2.2\% on Spider-Syn, Spider-Realistic, and Spider-DK, respectively, reaching the state-of-the-art performance. Particularly, the results from Spider-DK demonstrate that it significantly enhances the domain knowledge reasoning capability of fine-tuned LLMs.

\subsection{Ablation Study}
\begin{figure}[tb]
	\centering
	\small
	\includegraphics[width=0.75\columnwidth]{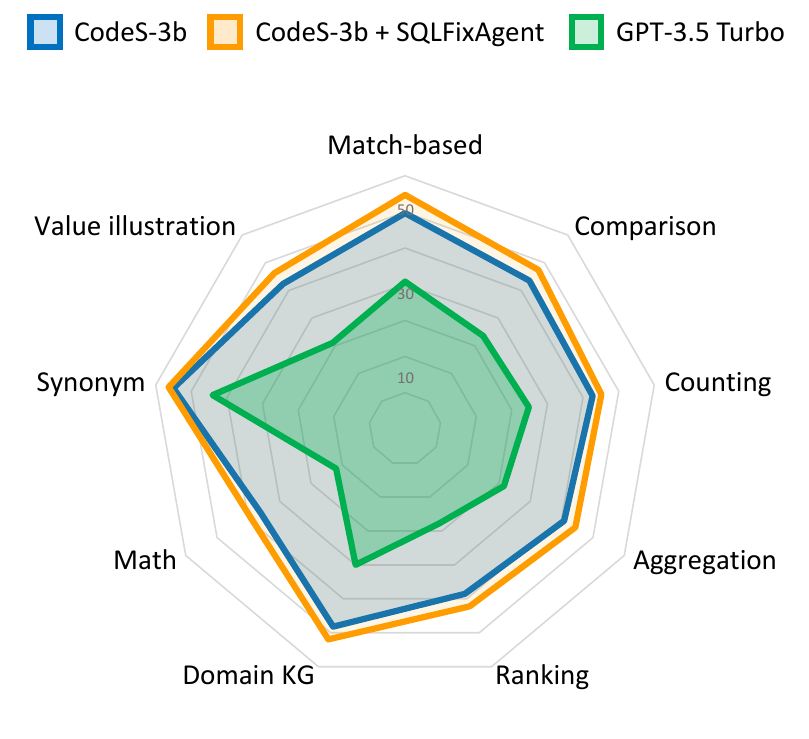}
	\caption{The fine-grained categorical evaluation of backbone models on BIRD’s dev set.}
	\label{fig:radar}
	\vspace{-2mm}
\end{figure}

\paragraph{Analysis of Backbone Model Capabilities}
Figure~\ref{fig:radar} illustrates the performance of the backbone models of SQLFixAgent across different sub-capability dimensions on Bird-dev. The results indicate that fine-tuned CodeS-3b outperforms GPT-3.5 Turbo in all areas, particularly in ranking, numerical computing (math), and value illustration, with accuracy of GPT-3.5 Turbo in math and ranking being below 30\%. It suggests that foundation LLMs struggle with deep data science tasks, which often require mathematical computations and rankings within the context of vague user queries. However, all models perform well in domain KG, synonym, and match-based areas, benefiting from language and reasoning abilities acquired during pre-training. Despite GPT-3.5 Turbo's inferior performance compared to fine-tuned CodeS-3b across all fine-grained categories, we employ it as agents for error detection and correction on LLM-generated SQL, further improving the query accuracy of the system.

\paragraph{Analysis of Performance on Sub-tasks}
Our framework includes two main subtasks in Text-to-SQL parsing: SQL error detection and correction, which are respectively handled by the SQLReviewer agent and the SQLRefiner agent. We employ F1 score to evaluate the performance of SQLReviewer in SQL error detection, and then calculate the repair success rate of SQLRefiner after the errors have been detected. As shown in Table~\ref{tab:subtask}, our framework can effectively detect erroneous SQL statements, including both semantic and syntactic errors. Due to the performance limitations of the backbone model, SQLFixAgent achieves repair success rates of 17.49\% and 15.64\% for errors generated by CodeS-3B and CodeS-7B, respectively. This presents a challenging task, and we plan to delve deeper into it in future work.

\begin{table}[tb]
	\centering
	\resizebox{\columnwidth}{!}{%
		\begin{tabular}{lcc}
			\toprule
			\multirow{2}{*}{\textbf{Methods}} & \multicolumn{2}{c}{\textbf{Dev set}} \\
			& \textbf{EX\%} & \textbf{VES\%} \\ \midrule
			\multicolumn{3}{c}{\cellcolor[HTML]{E6E6E6}\textbf{ICL-based}} \\ \midrule
			Codex~\citep{li2023llm} & 34.35 & 43.41 \\
			ChatGPT~\citep{li2023llm} & 37.22 & 43.81 \\
			GPT-4~\citep{li2023llm} & 46.35 & 49.77 \\
			DIN-SQL + GPT-4~\citep{pourreza2023dinsql} & 50.72 & 58.79 \\
			DAIL-SQL + GPT-4~\citep{gao2023texttosql} & 54.76 & 56.08 \\ \midrule
			\multicolumn{3}{c}{\cellcolor[HTML]{E6E6E6}\textbf{FT-based}} \\ \midrule
			T5-3B~\citep{li2023llm} & 23.34 & 25.57 \\
			CodeS-3B~\citep{li2024codes} & 55.02 & 56.54 \\
			CodeS-7B~\citep{li2024codes} & 57.17 & 58.80 \\ \midrule
			\multicolumn{3}{c}{\cellcolor[HTML]{E6E6E6}\textbf{Agent-based}} \\ \midrule
			MAC-SQL + GPT-3.5 Turbo~\citep{li2023llm} & 47.33 & 51.42 \\
			\rowcolor{blue!15!} CodeS-3B + SQLFixAgent (GPT-3.5 Turbo) & 58.67 ($\uparrow$3.65) & 62.19 ($\uparrow$5.65) \\
			\rowcolor{blue!15!} CodeS-7B + SQLFixAgent (GPT-3.5 Turbo) & \textbf{60.17} ($\uparrow$3.00) & \textbf{63.15} ($\uparrow$4.35) \\ \bottomrule
		\end{tabular}%
	}
	\caption{Evaluation of SQLFixAgent on BIRD’s dev set.}
	\label{tab:bird_result}
\end{table}

\paragraph{Analysis of ICL Methods} 
Table~\ref{tab:ICL_Ablation} presents the ablation study of the ICL-based methods used in SQLFixAgent. Starting from the complete framework, we successively removed rubber duck debugging (COT) from SQLReviewer, failure memory reflection (Reflexion), and multiple choice selection (MCS) from SQLRefiner and QueryCrafter. The experimental results show that each component of the SQLFixAgent framework is essential for generating semantically consistent results, as removing any of them led to decreased performance across all difficulty levels.

\begin{table*}[]
	\centering
	\resizebox{0.9\textwidth}{!}{%
		\begin{tabular}{c|lcccc}
			\toprule
			& \multirow{2}{*}{\textbf{Methods}} & \multicolumn{2}{c}{\textbf{Dev}} & \multicolumn{2}{c}{\textbf{Test}} \\
			&  & \textbf{EM\%} & \textbf{EX\%} & \textbf{EM\%} & \textbf{EX\%} \\ \midrule
			\multirow{5}{*}{\textbf{ICL-based}} & DAIL-SQL + GPT-4 + SC~\citep{gao2023texttosql} & 68.7 & 83.6 & 66.0 & 86.6 \\
			& DEA-SQL + GPT-4~\citep{xie2024decomposition} & - & 85.4 & - & \textbf{87.1} \\
			& C3 + ChatGPT + Zero-Shot~\citep{dong2023c3} & 71.4 & 81.8 & - & 82.3 \\
			& ChatGPT~\citep{liu2023comprehensive} & 34.6 & 74.4 & - & - \\
			& GPT-4~\citep{openai2024gpt4} & 22.1 & 72.3 & - & - \\ \midrule
			\multirow{6}{*}{\textbf{FT-based}} & T5-3B + PICARD~\citep{scholak2021picard} & 75.5 & 79.3 & 71.9 & 75.1 \\
			& Graphix-T5-3B + PICARD~\citep{li2023graphix} & 77.1 & 81.0 & \textbf{74.0} & 77.6 \\
			& RESDSQL-3B + NatSQL~\citep{li2023resdsql} & \textbf{80.5} & 84.1 & 72.0 & 79.9 \\
			& SQL-PaLM~\citep{sun2024sqlpalm} & 78.2 & 82.8 & - & - \\
			& SFT CodeS-3B~\citep{li2024codes} & 78.1 & 83.4 & 72.0 & 81.9 \\
			& SFT CodeS-7B~\citep{li2024codes} & 80.3 & 85.4 & 72.7 & 83.3 \\ \midrule
			\multirow{3}{*}{\textbf{Agent-based}} & MAC-SQL + GPT-3.5 Turbo~\citep{li2023llm} & 21.2 & 74.3 & 17.3 & 75.5 \\
			& \cellcolor{blue!15!}CodeS-3B + SQLFixAgent (GPT-3.5 Turbo) & \cellcolor{blue!15!}77.9 & \cellcolor{blue!15!}84.8 ($\uparrow$1.4) & \cellcolor{blue!15!}71.2 & \cellcolor{blue!15!}82.9 ($\uparrow$1.0) \\
			& \cellcolor{blue!15!}CodeS-7B + SQLFixAgent (GPT-3.5 Turbo) & \cellcolor{blue!15!}79.1 & \cellcolor{blue!15!}\textbf{86.2} ($\uparrow$0.8)& \cellcolor{blue!15!}72.6 & \cellcolor{blue!15!}83.9 ($\uparrow$0.6) \\ \bottomrule
		\end{tabular}%
	}
	\caption{Evaluation of SQLFixAgent on Spider’s dev/test sets.}
	\label{tab:spider_result}
\end{table*}

\begin{table*}[tb]
	\centering
	\resizebox{0.9\textwidth}{!}{%
		\begin{tabular}{lccccc}
			\toprule
			\multirow{2}{*}{\textbf{Methods}} & \multicolumn{2}{c}{\textbf{Spider-Syn}} & \multicolumn{2}{c}{\textbf{Spider-Realistic}} & \textbf{Spider-DK} \\
			& \textbf{EM\%} & \textbf{EX\%} & \textbf{EM\%} & \textbf{EX\%} & \textbf{EX\%} \\ \midrule
			T5-3B + PICARD~\citep{scholak2021picard} & 61.8 & 69.8 & 61.7 & 71.4 & 62.5 \\
			RESDSQL-3B + NatSQL~\citep{li2023resdsql} & 66.8 & 76.9 & 70.1 & 81.9 & 66.0 \\
			ChatGPT~\citep{liu2023comprehensive} & 48.5 & 58.6 & 49.2 & 63.4 & 62.6 \\
			SQL-Palm (Few-shot)~\citep{sun2024sqlpalm} & 67.4 & 74.6 & 72.4 & 77.6 & 66.5 \\
			SQL-Palm (Fine-tuned)~\citep{sun2024sqlpalm} & 66.4 & 70.9 & 73.2 & 77.4 & 67.5 \\ \midrule
			CodeS-7B~\citep{li2024codes} & \textbf{70.0} & 76.9 & \textbf{77.2} & 82.9 & 72.0 \\
			\rowcolor{blue!15!}CodeS-7B + SQLFixAgent (GPT-3.5 Turbo) & 68.4 & \textbf{77.7} ($\uparrow$0.8) & 75.8 & \textbf{83.3} ($\uparrow$0.4) & \textbf{74.2} ($\uparrow$2.2) \\ \bottomrule
		\end{tabular}%
	}
	\caption{Evaluation of SQLFixAgent on Spider variants.}
	\label{tab:variant_result}
\end{table*}

\begin{table}[tb]
	\centering
	\resizebox{0.9\columnwidth}{!}{%
		\begin{tabular}{lcc}
			\toprule
			\textbf{Methods} & \textbf{Detection (F1)} & \textbf{Repair (\%)} \\ \midrule
			CodeS-3B + SQLFixAgent & 73.73 & 17.49 \\
			CodeS-7B + SQLFixAgent & 73.68 & 15.64 \\ \bottomrule
		\end{tabular}%
	}
	\caption{Error detection accuracy and successful repair rate of SQLFixAgent on the Bird's dev set.}
	\label{tab:subtask}
\end{table}

\begin{figure}[tb]
	\centering
	\includegraphics[width=0.88\columnwidth]{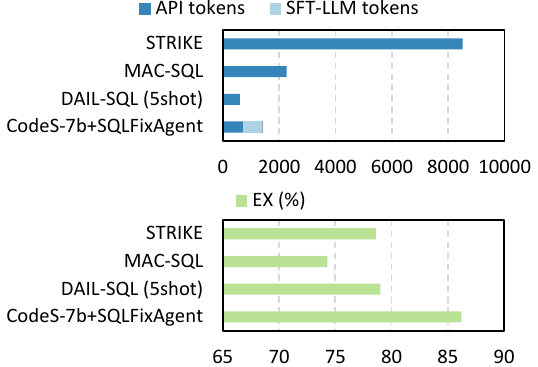}
	\caption{Token consumption and execution accuracy of advanced Text-to-SQL methods on Spider’s dev set.}
	\label{fig:token_efficiency}
\end{figure}

\begin{table}[tb]
	\resizebox{\columnwidth}{!}{%
		\begin{tabular}{lcccc}
			\toprule
			\textbf{Methods} & \textbf{Simple} & \textbf{Mod.} & \textbf{Chall.} & \textbf{All} \\ \midrule
			CodeS-3B + SQLFixAgent & 66.16 & 49.89 & 38.89 & 58.67  \\
			w/o COT & 65.73 & 48.82 & 39.58 & 58.15 ($\downarrow$)\\
			w/o Reflexion & 65.19 & 47.74 & 40.28 & 57.56 ($\downarrow$)\\
			w/o MCS & 64.86 & 47.53 & 38.89 & 57.17 ($\downarrow$)\\ \bottomrule
		\end{tabular}%
	}
	\caption{Execution accuracy of ICL method ablation study on BIRD’s dev set.}
	\label{tab:ICL_Ablation}
\end{table}

\subsection{Token Efficiency}
Considering that the LLM APIs charge based on token number, and the inference time is proportional to token lengths, we emphasize the token efficiency of Text-to-SQL system. In Figure 6, we demonstrate the average token consumption and execution accuracy of our method compare to three advanced methods, STRIKE~\citep{nan2023enhancing}, MAC-SQL and DAIL-SQL, they are all powered by GPT-3.5-turbo for a fair comparison. Since the SQLFixAgent initiates repairs only when the erroneous SQL is detected, it achieves state-of-art execution accuracy with fewer API token consumption, exhibiting higher token efficiency compared to other methods.

\section{Conclusion}
In this paper, we highlight the semantic mismatch error in LLM-based Text-to-SQL parsing and propose SQLFixAgent, a consistency-enhanced multi-agent collaborative framework designed for detecting and repairing the erroneous SQL generated by fine-tuned LLMs. The extensive experiments on Bird, Spider and its three robustness variants showcase the effectiveness and universality of our method. We validated our opinion through ablation experiments, where the inexperienced foundation LLMs can serve as agents to assist the fine-tuned LLMs in generating more semantic-accurate SQL queries. Furthermore, we underline the high token efficiency of SQLFixAgent compared to other advanced methods, providing direction for developing more practical Text-to-SQL solutions in real-world scenarios. In our future work, we would like to introduce more information (e.g., knowledge graph~\citep{jiang2024kgagent}) to further enhance the agents' capabilities in Text-to-SQL parsing. 
\newpage
\section{Acknowledgments}
We thank our anonymous reviewers for their helpful comments. This work was supported by three NSFC grants, i.e., No.62006166, No.62376178 and No.62072323. This work was also supported by Collaborative Innovation Center of Novel Software Technology and Industrialization and supported by a Project Funded by the Priority Academic Program Development of Jiangsu Higher Education Institutions (PAPD).

\bibliography{aaai25}

\begin{thebibliography}{43}
\providecommand{\natexlab}[1]{#1}

\bibitem[{AutoGPT-Team(2023)}]{autogpt2023}
AutoGPT-Team. 2023.
\newblock AutoGPT: build and use AI agents.

\bibitem[{Cao et~al.(2023)Cao, Zhang, Xu, Li, Ma, Chen, and Yu}]{cao2023astormer}
Cao, R.; Zhang, H.; Xu, H.; Li, J.; Ma, D.; Chen, L.; and Yu, K. 2023.
\newblock ASTormer: An AST Structure-aware Transformer Decoder for Text-to-SQL.
\newblock \emph{arXiv preprint arXiv:2310.18662}.

\bibitem[{Deng, Chen, and Zhang(2022)}]{deng2022recent}
Deng, N.; Chen, Y.; and Zhang, Y. 2022.
\newblock Recent Advances in Text-to-{SQL}: A Survey of What We Have and What We Expect.
\newblock In \emph{Proceedings of the 29th International Conference on Computational Linguistics}, 2166--2187. Gyeongju, Republic of Korea: International Committee on Computational Linguistics.

\bibitem[{Deng et~al.(2021)Deng, Awadallah, Meek, Polozov, Sun, and Richardson}]{Deng_2021}
Deng, X.; Awadallah, A.~H.; Meek, C.; Polozov, O.; Sun, H.; and Richardson, M. 2021.
\newblock Structure-Grounded Pretraining for Text-to-{SQL}.
\newblock In \emph{Proceedings of the 2021 Conference of the North American Chapter of the Association for Computational Linguistics: Human Language Technologies}, 1337--1350. Online: Association for Computational Linguistics.

\bibitem[{Dong et~al.(2023)Dong, Zhang, Ge, Mao, Gao, lu~Chen, Lin, and Lou}]{dong2023c3}
Dong, X.; Zhang, C.; Ge, Y.; Mao, Y.; Gao, Y.; lu~Chen; Lin, J.; and Lou, D. 2023.
\newblock C3: Zero-shot Text-to-SQL with ChatGPT.
\newblock arXiv:2307.07306.

\bibitem[{Durante et~al.(2024)Durante, Huang, Wake, Gong, Park, Sarkar, Taori, Noda, Terzopoulos, and Choi}]{durante2024agent}
Durante, Z.; Huang, Q.; Wake, N.; Gong, R.; Park, J.~S.; Sarkar, B.; Taori, R.; Noda, Y.; Terzopoulos, D.; and Choi, Y. 2024.
\newblock Agent AI: Surveying the Horizons of Multimodal Interaction.
\newblock arXiv:2401.03568.

\bibitem[{Gan et~al.(2021)Gan, Chen, Huang, Purver, Woodward, Xie, and Huang}]{gan2021robustness}
Gan, Y.; Chen, X.; Huang, Q.; Purver, M.; Woodward, J.~R.; Xie, J.; and Huang, P. 2021.
\newblock Towards Robustness of Text-to-{SQL} Models against Synonym Substitution.
\newblock In \emph{Proceedings of the 59th Annual Meeting of the Association for Computational Linguistics and the 11th International Joint Conference on Natural Language Processing (Volume 1: Long Papers)}, 2505--2515. Online: Association for Computational Linguistics.

\bibitem[{Gan, Chen, and Purver(2021)}]{gan2021exploring}
Gan, Y.; Chen, X.; and Purver, M. 2021.
\newblock Exploring Underexplored Limitations of Cross-Domain Text-to-{SQL} Generalization.
\newblock In \emph{Proceedings of the 2021 Conference on Empirical Methods in Natural Language Processing}, 8926--8931. Online and Punta Cana, Dominican Republic: Association for Computational Linguistics.

\bibitem[{Gao et~al.(2024)Gao, Wang, Li, Sun, Qian, Ding, and Zhou}]{gao2023texttosql}
Gao, D.; Wang, H.; Li, Y.; Sun, X.; Qian, Y.; Ding, B.; and Zhou, J. 2024.
\newblock Text-to-SQL Empowered by Large Language Models: A Benchmark Evaluation.
\newblock \emph{Proc. VLDB Endow.}, 17(5): 1132–1145.

\bibitem[{Guo et~al.(2019)Guo, Zhan, Gao, Xiao, Lou, Liu, and Zhang}]{guo-etal-2019-towards}
Guo, J.; Zhan, Z.; Gao, Y.; Xiao, Y.; Lou, J.-G.; Liu, T.; and Zhang, D. 2019.
\newblock Towards Complex Text-to-{SQL} in Cross-Domain Database with Intermediate Representation.
\newblock In \emph{Proceedings of the 57th Annual Meeting of the Association for Computational Linguistics}, 4524--4535. Florence, Italy: Association for Computational Linguistics.

\bibitem[{Hong et~al.(2023)Hong, Zhuge, Chen, Zheng, Cheng, Zhang, Wang, Wang, Yau, Lin, and Zhou}]{hong2023metagpt}
Hong, S.; Zhuge, M.; Chen, J.; Zheng, X.; Cheng, Y.; Zhang, C.; Wang, J.; Wang, Z.; Yau, S. K.~S.; Lin, Z.; and Zhou, L. 2023.
\newblock MetaGPT: Meta Programming for A Multi-Agent Collaborative Framework.
\newblock arXiv:2308.00352.

\bibitem[{Huang et~al.(2023)Huang, Yu, Ma, Zhong, Feng, Wang, Chen, Peng, Feng, Qin, and Liu}]{huang2023survey}
Huang, L.; Yu, W.; Ma, W.; Zhong, W.; Feng, Z.; Wang, H.; Chen, Q.; Peng, W.; Feng, X.; Qin, B.; and Liu, T. 2023.
\newblock A Survey on Hallucination in Large Language Models: Principles, Taxonomy, Challenges, and Open Questions.
\newblock arXiv:2311.05232.

\bibitem[{Hui et~al.(2022)Hui, Geng, Wang, Qin, Li, Li, Sun, and Li}]{hui-etal-2022-s2sql}
Hui, B.; Geng, R.; Wang, L.; Qin, B.; Li, Y.; Li, B.; Sun, J.; and Li, Y. 2022.
\newblock {S}$^2${SQL}: Injecting Syntax to Question-Schema Interaction Graph Encoder for Text-to-{SQL} Parsers.
\newblock In \emph{Findings of the Association for Computational Linguistics: ACL 2022}, 1254--1262. Dublin, Ireland: Association for Computational Linguistics.

\bibitem[{Jiang et~al.(2023)Jiang, Zhou, Dong, Ye, Zhao, and Wen}]{jiang2023structgpt}
Jiang, J.; Zhou, K.; Dong, Z.; Ye, K.; Zhao, X.; and Wen, J.-R. 2023.
\newblock {S}truct{GPT}: A General Framework for Large Language Model to Reason over Structured Data.
\newblock In \emph{Proceedings of the 2023 Conference on Empirical Methods in Natural Language Processing}, 9237--9251. Singapore: Association for Computational Linguistics.

\bibitem[{Jiang et~al.(2024)Jiang, Zhou, Zhao, Song, Zhu, Zhu, and Wen}]{jiang2024kgagent}
Jiang, J.; Zhou, K.; Zhao, W.~X.; Song, Y.; Zhu, C.; Zhu, H.; and Wen, J.-R. 2024.
\newblock KG-Agent: An Efficient Autonomous Agent Framework for Complex Reasoning over Knowledge Graph.
\newblock arXiv:2402.11163.

\bibitem[{Lee et~al.(2024)Lee, Xia, tse Huang, Zhu, Zhang, and Lyu}]{lee2024unified}
Lee, C.; Xia, C.~S.; tse Huang, J.; Zhu, Z.; Zhang, L.; and Lyu, M.~R. 2024.
\newblock A Unified Debugging Approach via LLM-Based Multi-Agent Synergy.
\newblock arXiv:2404.17153.

\bibitem[{Li et~al.(2023{\natexlab{a}})Li, Zhang, Li, and Chen}]{li2023resdsql}
Li, H.; Zhang, J.; Li, C.; and Chen, H. 2023{\natexlab{a}}.
\newblock Resdsql: Decoupling schema linking and skeleton parsing for text-to-sql.
\newblock In \emph{Proceedings of the AAAI Conference on Artificial Intelligence}, 13067--13075.

\bibitem[{Li et~al.(2024{\natexlab{a}})Li, Zhang, Liu, Fan, Zhang, Zhu, Wei, Pan, Li, and Chen}]{li2024codes}
Li, H.; Zhang, J.; Liu, H.; Fan, J.; Zhang, X.; Zhu, J.; Wei, R.; Pan, H.; Li, C.; and Chen, H. 2024{\natexlab{a}}.
\newblock CodeS: Towards Building Open-source Language Models for Text-to-SQL.
\newblock \emph{Proc. ACM Manag. Data}, 2(3).

\bibitem[{Li et~al.(2023{\natexlab{b}})Li, Hui, Cheng, Qin, Ma, Huo, Huang, Du, Si, and Li}]{li2023graphix}
Li, J.; Hui, B.; Cheng, R.; Qin, B.; Ma, C.; Huo, N.; Huang, F.; Du, W.; Si, L.; and Li, Y. 2023{\natexlab{b}}.
\newblock Graphix-t5: Mixing pre-trained transformers with graph-aware layers for text-to-sql parsing.
\newblock In \emph{Proceedings of the AAAI Conference on Artificial Intelligence}, 13076--13084.

\bibitem[{Li et~al.(2024{\natexlab{b}})Li, Hui, Qu, Yang, Li, Li, Wang, Qin, Geng, Huo, Zhou, Ma, and Li}]{li2023llm}
Li, J.; Hui, B.; Qu, G.; Yang, J.; Li, B.; Li, B.; Wang, B.; Qin, B.; Geng, R.; Huo, N.; Zhou, X.; Ma, C.; and Li, G. 2024{\natexlab{b}}.
\newblock Can LLM already serve as a database interface? a big bench for large-scale database grounded text-to-SQLs.
\newblock In \emph{Proceedings of the 37th International Conference on Neural Information Processing Systems}. Red Hook, NY, USA: Curran Associates Inc.

\bibitem[{Li et~al.(2023{\natexlab{c}})Li, Allal, Zi, Muennighoff, Kocetkov, Mou, Marone, Akiki, Li, and Chim}]{li2023starcoder}
Li, R.; Allal, L.~B.; Zi, Y.; Muennighoff, N.; Kocetkov, D.; Mou, C.; Marone, M.; Akiki, C.; Li, J.; and Chim, J. 2023{\natexlab{c}}.
\newblock StarCoder: may the source be with you!
\newblock arXiv:2305.06161.

\bibitem[{Liu et~al.(2023)Liu, Hu, Wen, and Yu}]{liu2023comprehensive}
Liu, A.; Hu, X.; Wen, L.; and Yu, P.~S. 2023.
\newblock A comprehensive evaluation of ChatGPT's zero-shot Text-to-SQL capability.
\newblock arXiv:2303.13547.

\bibitem[{Nan et~al.(2023)Nan, Zhao, Zou, Ri, Tae, Zhang, Cohan, and Radev}]{nan2023enhancing}
Nan, L.; Zhao, Y.; Zou, W.; Ri, N.; Tae, J.; Zhang, E.; Cohan, A.; and Radev, D. 2023.
\newblock Enhancing Text-to-{SQL} Capabilities of Large Language Models: A Study on Prompt Design Strategies.
\newblock In \emph{Findings of the Association for Computational Linguistics: EMNLP 2023}, 14935--14956. Singapore: Association for Computational Linguistics.

\bibitem[{OpenAI et~al.(2024)OpenAI, Achiam, Adler, Agarwal, Ahmad, Akkaya, Aleman, Almeida, and Altenschmidt}]{openai2024gpt4}
OpenAI; Achiam, J.; Adler, S.; Agarwal, S.; Ahmad, L.; Akkaya, I.; Aleman, F.~L.; Almeida, D.; and Altenschmidt, J. 2024.
\newblock GPT-4 Technical Report.
\newblock arXiv:2303.08774.

\bibitem[{Pourreza and Rafiei(2023)}]{pourreza2023dinsql}
Pourreza, M.; and Rafiei, D. 2023.
\newblock DIN-SQL: Decomposed In-Context Learning of Text-to-SQL with Self-Correction.
\newblock arXiv:2304.11015.

\bibitem[{Qin et~al.(2022)Qin, Hui, Wang, Yang, Li, Li, Geng, Cao, Sun, Si, Huang, and Li}]{qin2022survey}
Qin, B.; Hui, B.; Wang, L.; Yang, M.; Li, J.; Li, B.; Geng, R.; Cao, R.; Sun, J.; Si, L.; Huang, F.; and Li, Y. 2022.
\newblock A Survey on Text-to-SQL Parsing: Concepts, Methods, and Future Directions.
\newblock arXiv:2208.13629.

\bibitem[{Scholak, Schucher, and Bahdanau(2021)}]{scholak2021picard}
Scholak, T.; Schucher, N.; and Bahdanau, D. 2021.
\newblock {PICARD}: Parsing Incrementally for Constrained Auto-Regressive Decoding from Language Models.
\newblock In \emph{Proceedings of the 2021 Conference on Empirical Methods in Natural Language Processing}, 9895--9901. Online and Punta Cana, Dominican Republic: Association for Computational Linguistics.

\bibitem[{Shinn et~al.(2023)Shinn, Cassano, Gopinath, Narasimhan, and Yao}]{shinn2023reflexion}
Shinn, N.; Cassano, F.; Gopinath, A.; Narasimhan, K.; and Yao, S. 2023.
\newblock Reflexion: language agents with verbal reinforcement learning.
\newblock In \emph{Advances in Neural Information Processing Systems}, 8634--8652. Curran Associates, Inc.

\bibitem[{Sun et~al.(2024)Sun, Arik, Muzio, Miculicich, Gundabathula, Yin, Dai, Nakhost, and Sinha}]{sun2024sqlpalm}
Sun, R.; Arik, S.~{\"O}.; Muzio, A.; Miculicich, L.; Gundabathula, S.; Yin, P.; Dai, H.; Nakhost, H.; and Sinha, R. 2024.
\newblock SQL-PaLM: Improved Large Language Model Adaptation for Text-to-SQL (extended).
\newblock arXiv:2306.00739.

\bibitem[{Sun et~al.(2023)Sun, Zhang, Yan, Gao, and Ong}]{sun2023battle}
Sun, S.; Zhang, Y.; Yan, J.; Gao, Y.; and Ong, D. 2023.
\newblock Battle of the Large Language Models: Dolly vs {LL}a{MA} vs Vicuna vs Guanaco vs Bard vs {C}hat{GPT} - A Text-to-{SQL} Parsing Comparison.
\newblock In \emph{Findings of the Association for Computational Linguistics: EMNLP 2023}, 11225--11238. Singapore: Association for Computational Linguistics.

\bibitem[{Team et~al.(2024)Team, Anil, Borgeaud, Alayrac, Yu, Soricut, Schalkwyk, Dai, Hauth, and Millican}]{geminiteam2024gemini}
Team, G.; Anil, R.; Borgeaud, S.; Alayrac, J.-B.; Yu, J.; Soricut, R.; Schalkwyk, J.; Dai, A.~M.; Hauth, A.; and Millican, K. 2024.
\newblock Gemini: A Family of Highly Capable Multimodal Models.
\newblock arXiv:2312.11805.

\bibitem[{Touvron et~al.(2023)Touvron, Lavril, Izacard, Martinet, Lachaux, Lacroix, Rozière, Goyal, and Hambro}]{touvron2023llama}
Touvron, H.; Lavril, T.; Izacard, G.; Martinet, X.; Lachaux, M.-A.; Lacroix, T.; Rozière, B.; Goyal, N.; and Hambro, E. 2023.
\newblock LLaMA: Open and Efficient Foundation Language Models.
\newblock arXiv:2302.13971.

\bibitem[{Wang et~al.(2024{\natexlab{a}})Wang, Ren, Yang, Liang, Bai, Chai, Yan, Zhang, Yin, Sun, and Li}]{wang2024macsql}
Wang, B.; Ren, C.; Yang, J.; Liang, X.; Bai, J.; Chai, L.; Yan, Z.; Zhang, Q.-W.; Yin, D.; Sun, X.; and Li, Z. 2024{\natexlab{a}}.
\newblock MAC-SQL: A Multi-Agent Collaborative Framework for Text-to-SQL.
\newblock arXiv:2312.11242.

\bibitem[{Wang et~al.(2020)Wang, Shin, Liu, Polozov, and Richardson}]{wang2019rat}
Wang, B.; Shin, R.; Liu, X.; Polozov, O.; and Richardson, M. 2020.
\newblock {RAT-SQL}: Relation-Aware Schema Encoding and Linking for Text-to-{SQL} Parsers.
\newblock In \emph{Proceedings of the 58th Annual Meeting of the Association for Computational Linguistics}, 7567--7578. Online: Association for Computational Linguistics.

\bibitem[{Wang et~al.(2024{\natexlab{b}})Wang, Ma, Feng, Zhang, Yang, Zhang, Chen, Tang, Chen, Lin, Zhao, Wei, and Wen}]{Wang_2024}
Wang, L.; Ma, C.; Feng, X.; Zhang, Z.; Yang, H.; Zhang, J.; Chen, Z.; Tang, J.; Chen, X.; Lin, Y.; Zhao, W.~X.; Wei, Z.; and Wen, J. 2024{\natexlab{b}}.
\newblock A survey on large language model based autonomous agents.
\newblock \emph{Frontiers of Computer Science}, 18(6).

\bibitem[{Wei et~al.(2022{\natexlab{a}})Wei, Wang, Schuurmans, Bosma, brian ichter, Xia, Chi, Le, and Zhou}]{wei2022chain}
Wei, J.; Wang, X.; Schuurmans, D.; Bosma, M.; brian ichter; Xia, F.; Chi, E.~H.; Le, Q.~V.; and Zhou, D. 2022{\natexlab{a}}.
\newblock Chain of Thought Prompting Elicits Reasoning in Large Language Models.
\newblock In Oh, A.~H.; Agarwal, A.; Belgrave, D.; and Cho, K., eds., \emph{Advances in Neural Information Processing Systems}.

\bibitem[{Wei et~al.(2022{\natexlab{b}})Wei, Wang, Schuurmans, Bosma, ichter, Xia, Chi, Le, and Zhou}]{NEURIPS2022_9d560961}
Wei, J.; Wang, X.; Schuurmans, D.; Bosma, M.; ichter, b.; Xia, F.; Chi, E.; Le, Q.~V.; and Zhou, D. 2022{\natexlab{b}}.
\newblock Chain-of-Thought Prompting Elicits Reasoning in Large Language Models.
\newblock In \emph{Advances in Neural Information Processing Systems}, 24824--24837. Curran Associates, Inc.

\bibitem[{Xi et~al.(2023)Xi, Chen, Guo, He, Ding, Hong, Zhang, Wang, Jin, Zhou, Zheng, and Fan}]{xi2023rise}
Xi, Z.; Chen, W.; Guo, X.; He, W.; Ding, Y.; Hong, B.; Zhang, M.; Wang, J.; Jin, S.; Zhou, E.; Zheng, R.; and Fan, X. 2023.
\newblock The Rise and Potential of Large Language Model Based Agents: A Survey.
\newblock arXiv:2309.07864.

\bibitem[{Xie et~al.(2024)Xie, Jin, Xie, Lin, Chen, Yu, Cheng, Zhuo, Hu, and Li}]{xie2024decomposition}
Xie, Y.; Jin, X.; Xie, T.; Lin, M.; Chen, L.; Yu, C.; Cheng, L.; Zhuo, C.; Hu, B.; and Li, Z. 2024.
\newblock Decomposition for Enhancing Attention: Improving LLM-based Text-to-SQL through Workflow Paradigm.
\newblock \emph{arXiv preprint arXiv:2402.10671}.

\bibitem[{Yu et~al.(2018)Yu, Zhang, Yang, Yasunaga, Wang, Li, Ma, Li, Yao, Roman, and Zhang}]{yu2019spider}
Yu, T.; Zhang, R.; Yang, K.; Yasunaga, M.; Wang, D.; Li, Z.; Ma, J.; Li, I.; Yao, Q.; Roman, S.; and Zhang, Z. 2018.
\newblock {S}pider: A Large-Scale Human-Labeled Dataset for Complex and Cross-Domain Semantic Parsing and Text-to-{SQL} Task.
\newblock In \emph{Proceedings of the 2018 Conference on Empirical Methods in Natural Language Processing}, 3911--3921. Brussels, Belgium: Association for Computational Linguistics.

\bibitem[{Yu et~al.(2019)Yu, Zhang, Yasunaga, Tan, Lin, Li, Er, Li, Pang, Chen, Ji, and Dixit}]{yu-etal-2019-sparc}
Yu, T.; Zhang, R.; Yasunaga, M.; Tan, Y.~C.; Lin, X.~V.; Li, S.; Er, H.; Li, I.; Pang, B.; Chen, T.; Ji, E.; and Dixit, S. 2019.
\newblock {SP}ar{C}: Cross-Domain Semantic Parsing in Context.
\newblock In \emph{Proceedings of the 57th Annual Meeting of the Association for Computational Linguistics}, 4511--4523. Florence, Italy: Association for Computational Linguistics.

\bibitem[{Zhong, Yu, and Klein(2020)}]{ruiqi20}
Zhong, R.; Yu, T.; and Klein, D. 2020.
\newblock Semantic Evaluation for Text-to-SQL with Distilled Test Suite.
\newblock In \emph{The 2020 Conference on Empirical Methods in Natural Language Processing}. Association for Computational Linguistics.

\bibitem[{Zhou et~al.(2023)Zhou, Sch{\"a}rli, Hou, Wei, Scales, Wang, Schuurmans, Cui, Bousquet, Le, and Chi}]{zhou2023leasttomost}
Zhou, D.; Sch{\"a}rli, N.; Hou, L.; Wei, J.; Scales, N.; Wang, X.; Schuurmans, D.; Cui, C.; Bousquet, O.; Le, Q.~V.; and Chi, E.~H. 2023.
\newblock Least-to-Most Prompting Enables Complex Reasoning in Large Language Models.
\newblock In \emph{The Eleventh International Conference on Learning Representations}.

\end{thebibliography}

\end{document}